\documentclass[10pt,twocolumn,letterpaper]{article}

\usepackage{cvpr}              
\usepackage{graphicx}
\usepackage{amsmath}
\usepackage{amssymb}
\usepackage{booktabs}

\usepackage{makecell}
\usepackage{array}
\usepackage{algorithm}
\usepackage{algorithmic}
\usepackage[pagebackref,breaklinks,colorlinks]{hyperref}

\usepackage[capitalize]{cleveref}
\crefname{section}{Sec.}{Secs.}
\Crefname{section}{Section}{Sections}
\Crefname{table}{Table}{Tables}
\crefname{table}{Tab.}{Tabs.}

\usepackage{newfloat}
\usepackage{listings}
\usepackage{multirow}
\usepackage{tablefootnote}
\usepackage{enumitem}

\usepackage{xcolor}

\definecolor{codegreen}{rgb}{0.5,0.6,0.6}
\definecolor{codegray}{rgb}{0.5,0.5,0.5}
\definecolor{codepurple}{rgb}{0.58,0,0.82}
\definecolor{backcolour}{rgb}{0.95,0.95,0.92}

\lstdefinestyle{mystyle}{
  commentstyle=\color{codegreen},
  keywordstyle=\color{magenta},
  numberstyle=\tiny\color{codegray},
  stringstyle=\color{codepurple},
  basicstyle=\linespread{0.8}\ttfamily\scriptsize,
  breakatwhitespace=false,         
  breaklines=true,                 
  captionpos=b,                    
  keepspaces=false,  
  showspaces=false,                
  showstringspaces=false,
  showtabs=false,                  
  tabsize=1
}
\newsavebox{\mycode}

\def\cvprPaperID{*****} 
\def\confName{CVPR}
\def\confYear{2022}

\begin{document}

\title{Directional Self-supervised Learning for Heavy Image Augmentations}

\author{Yalong Bai\textsuperscript{1}\footnotemark[1]~~~~~~~~Yifan Yang\textsuperscript{2}\thanks{Equal contribution.}~~~~~~~Wei Zhang\textsuperscript{1}\thanks{Corresponding author.}~~~~~~~~Tao Mei\textsuperscript{1}\\
{\normalsize \textsuperscript{1}JD AI Research~~~~~~~~\textsuperscript{2}Peking University}\\
{\tt\small ylbai@outlook.com,  yif\_yang@pku.edu.cn,  wzhang.cu@gmail.com,  tmei@live.com}
}

\maketitle

\begin{abstract}
Despite the large augmentation family, only a few cherry-picked robust augmentation policies are beneficial to self-supervised image representation learning. 
In this paper, we propose a directional self-supervised learning paradigm (DSSL), which is compatible with significantly more augmentations. 
Specifically, we adapt heavy augmentation policies after the views lightly augmented by standard augmentations, to generate harder view (HV). HV usually has a higher deviation from the original image than the lightly augmented standard view (SV). Unlike previous methods equally pairing all augmented views to symmetrically maximize their similarities, DSSL treats augmented views of the same instance as a partially ordered set (with directions as SV$\leftrightarrow $SV, SV$\leftarrow$HV), and then equips a directional objective function respecting to the derived relationships among views. DSSL can be easily implemented with a few lines of codes and is highly flexible to popular self-supervised learning frameworks, including SimCLR, SimSiam, BYOL. Extensive experimental results on CIFAR and ImageNet demonstrated that DSSL can stably improve various baselines with compatibility to a wider range of augmentations.
\end{abstract}


\section{Introduction}
Unsupervised visual representation learning aims at learning image features without using manual semantic annotations. Recently, self-supervised learning driven by instance discrimination tasks~\cite{wu2018unsupervised,ye2019unsupervised,tian2019contrastive,he2020momentum,chen2020simple} or Siamese architecture~\cite{chen2020exploring,grill2020bootstrap} has achieved great success in learning high-quality visual features and closing the performance gap with supervised pretraining on various computer tasks. The visual embedding space of self-supervised learning method is constructed by minimizing the dissimilarity among representations of variations derived from the same image, and/or increasing the distance between the representations of augmented view from different images (negative pair). Image transformation, which aims to generate variations from the same image~\cite{chen2020simple,grill2020bootstrap,chen2020exploring,he2020momentum}, plays a crucial role in the self-supervised visual representations learning. 

\begin{figure}[t]
\begin{minipage}[c]{0.55\linewidth}
    \raggedright
    \footnotesize
    \setlength{\tabcolsep}{1.7mm}{
    \scriptsize
    \renewcommand{\arraystretch}{1.2}
    \begin{tabular}[c]{l|c|c}
  Augment & SimSiam & w/ DSSL \\
  \Xhline{2\arrayrulewidth}
\textit{Standard Aug.} & 92.17 &   \\\hline
w/ \textit{JigSaw}(2) & 92.79  & \textbf{93.56} \\
w/ \textit{JigSaw}(4) & 89.72 & \textbf{91.95} \\
\hline
w/ \textit{RA}(1,1) & 88.00 & \textbf{92.29} \\
w/ \textit{RA}(2,1) & 82.80 & \textbf{93.09} \\
w/ \textit{RA}(2,5) & 9.78 & \textbf{94.17} \\\hline
w/ \textit{UA} & 91.11 & \textbf{93.27} \\
\end{tabular}}
\hfill
\end{minipage}
  \begin{minipage}[c]{0.44\linewidth}
    \raggedleft
    \includegraphics[width=1\linewidth,page=1]{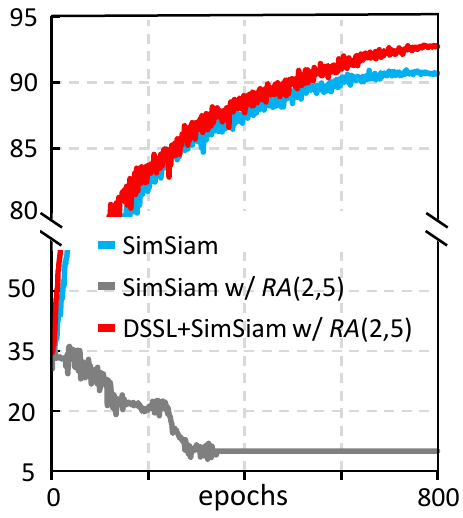}
  \end{minipage}
  
\caption{Left: Linear evaluation accuracy of SimSiam~\cite{chen2020exploring} on CIFAR-10 by adding extra \textit{heavy augmentations} besides the original \textit{standard augmentations}. The number of grids for \textit{JigSaw}($n$) is $n\times n$. \textit{RA}($m,n$) is the RandAugment~\cite{cubuk2020randaugment} with $n$ augmentation transformation of $m$ magnitude. \textit{UA} denotes the UniformAumgent~\cite{lingchen2020uniformaugment}. Right: validation accuracy of kNN classification during pre-training. Incorporating heavy augmentations on SimSiam results in unstable performance even \textit{collapsing} (linear evaluation on collapsed model tends to random guess). Our DSSL consistently benefits from heavy augmentations for higher performances.}
\label{fig:exp_motivation}
\end{figure}

\begin{figure*}[!t]
  \centering
  \includegraphics[width=0.97\linewidth,page=1]{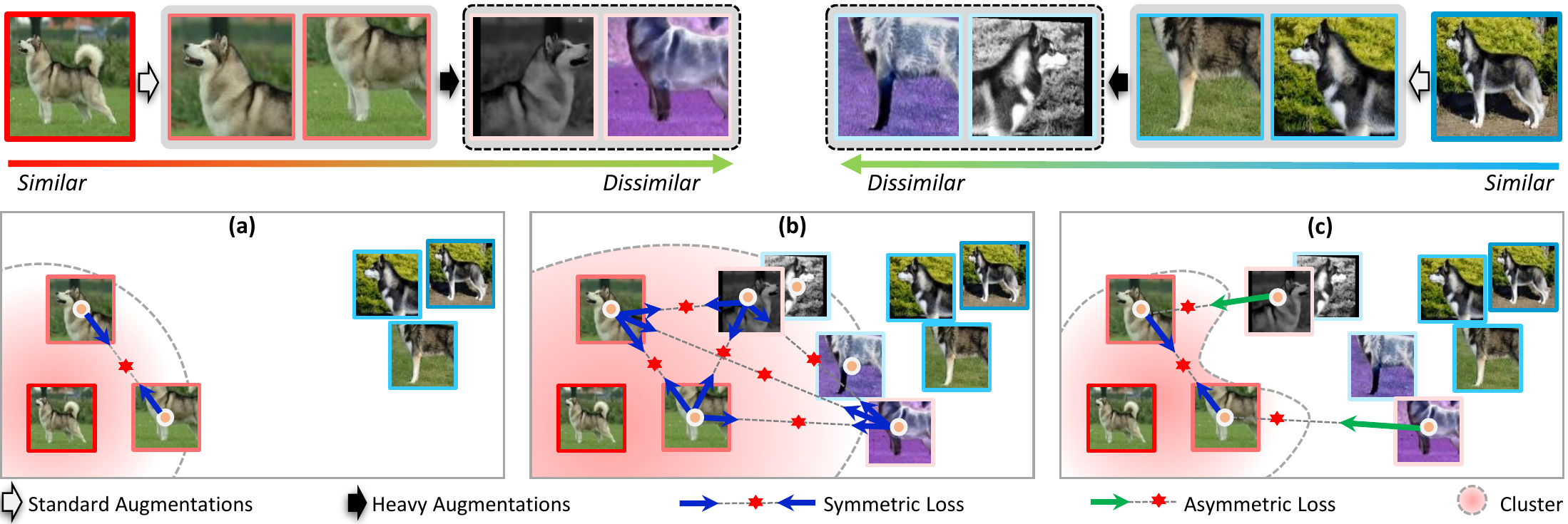}
  \caption{
  Overview of standard self-supervised learning and our DSSL. Original standard image transformations generate the standard views, and the harder view is derived from the the standard view by applying heavy augmentation \textit{RandAugment}. (a) Standard instance-wise learning with standard views. (b) Instance-wise self-supervised learning after introducing heavily augmented (harder) views. Applying symmetric loss to maximize the similarity between standard and heavily augmented views roughly expands the feature cluster in the visual embedding space. The model may confuse the instance-level identity. (c) DSSL: To prevent the adverse effect from \textit{missing information} of heavily augmented views, DSSL avoids arbitrarily maximizing their visual agreement. To tighten the feature cluster, DSSL applies an asymmetric loss for only gathering each heavily augmented view to its relevant standard view.}
  \label{fig:motivation}
\end{figure*}

However, these self-supervised learning methods share a common fundamental weakness: only a few carefully selected augmentation policies with proper settings are beneficial to the model training. The combination of \textit{random crop, color distortion, Gaussian blur and grayscale} is crucial to achieving good performance, applied as the basic augmentation setting for many popular instance-wise self-supervised learning methods~\cite{caron2020unsupervised,he2020momentum,chen2020exploring,li2020prototypical,grill2020bootstrap}. Here we define these augmentations as the \textbf{standard augmentations}. We denote the images augmented by standard augmentations as the standard views. Recent works find that a few other augmentations (e.g., \textit{RandAugment}, \textit{JigSaw}) can further improve the performance of self-supervised learning methods rely on negative pairs~\cite{tian2020makes}. However, for negative pair free self-supervised learning methods like SimSiam~\cite{chen2020exploring}, introducing such augmentations usually results in a lousy performance, even model collapsing during training, as shown in \cref{fig:exp_motivation}. We name these unstable and risky data augmentation policies as \textbf{heavy augmentations}, since they usually largely alters the image appearance.

Inspired by previous works~\cite{li2020prototypical,chen2020exploring} that formulating the instance-wise self-supervised learning as K-means clustering of all augmented views from the same instance, we hypothesize a gold standard feature cluster for all views of one given image instance existing in the visual feature embedding space, and define $d$ as the deviation of augmented view's feature from the core-point of its relevant gold standard feature cluster. Standard self-supervised learning methods treat all augmented views of the same image fairly to construct training pairs. As shown in \cref{fig:motivation} (a), such a strategy works well, and the model can converge stably for the standard image transformations. However, after incorporating the views augmented from heavy image transformations (\cref{fig:motivation} (b)), two obvious risks arise. 1) Closing the representation of standard views to heavily augmented views would roughly expand the feature cluster in the embedding space. This would increase the difficulty of constructing an embedding space where all instances are well-separated and also may cause unexpected confusion for instance-wise discriminating, especially for those negative pairs free methods. 2) Maximizing the visual agreement among heavily augmented views disaccords with the ``InfoMin principle''~\cite{tian2020makes}. Since the mutual information among views with large $d$ is usually low, contrasting these views leads to \textit{missing information} and results in poor performance in downstream tasks.

To address this, we propose Directional Self-supervised Learning (DSSL), a new training method for unsupervised representation learning that could stably improve the performance of instance-wise self-supervised learning by completely applying more heavy image transformations. Fig.~\ref{fig:motivation} (c) shows an illustration of DSSL. For each standard view (SV) augmented from the original robust image transformations, we can generate various harder views (HV) derived from it by applying additional heavy augmentation policies. These heavily augmented view has a larger $d$ than its relevant standard view. In this way, we can treat all augmented views of the same image as a partially ordered set (SV$\leftrightarrow $SV, SV$\leftarrow$HV) in terms of $d$.
An asymmetric loss is introduced to encourage the representation of each heavily augmented view (HV) to be close to its relevant source standard view (SV). In this way, the feature cluster for all augmented views can be presented as non-convex, rather than the K-means convex clustering, the whole cluster is tightened. Moreover, DSSL discards the instance-wise self-supervised learning among RVs to bypass the issue of low mutual information among HVs. As a result, more augmentation policies can be introduced to enrich the information of the whole embedding space but keep the instances still well-separated.

DSSL is a straightforward algorithm that can be easily implemented in a few line of Pseudo code. Also, there are no additional hyper-parameter needs to be adjusted in DSSL. 
We validate the effectiveness of DSSL by evaluating it on several self-supervised benchmarks. In particular, on the ImageNet linear evaluation protocol, DSSL achieved stable performance improvements. All DSSL based pre-training models surpass the supervised pre-training model on the CIFAR-10 linear evaluation. Moreover, the transfer performance on detection and segmentation task further demonstrate the efficiency of DSSL with heavy augmentations. 

The main contributions are summarized as follows:

\begin{itemize}[itemsep=2pt,topsep=0pt,parsep=0pt]
	\item A novel Directional Self-supervised Learning (DSSL) paradigm is proposed for unsupervised visual representation learning. We introduce a partially ordered set to organize the augmented views, and introduce an asymmetric loss for harnessing rich information from heavy augmented views.
	\item DSSL is easy-to-implement and applicable to various standard instance-wise self-supervised learning frameworks by introducing minor modifications without any hyper-parameters.
	
	\item DSSL stably improves over various self-supervised learning methods on standard benchmarks, even when thoroughly applying heavy image transformations that show adverse effect to previous methods.
\end{itemize}
\section{Related Work}

Our work is related to studies in the 
instance-wise self-supervised learning and data augmentation policies.


\noindent\textbf{Instance-wise self-supervised learning.}\quad Instance-level classification task treats each image and its variants as one specific class. It aims to construct visual embedding space by pull all samples in the same class close while pushing samples from other classes away. Since it is hard to directly categorize all training samples into a large number of classes~\cite{dosovitskiy2015discriminative}, the early instance-wise contrastive learning method replaces the classifier with a memory bank~\cite{wu2018unsupervised} to store the previous features of all samples calculated in the previous stage and sample positive and negative pairs from the memory bank. Several other technologies have also been adopted and extended based on this method, such as introducing local similarity~\cite{zhuang2019local} and neighborhood discovery~\cite{huang2019unsupervised} for further improving the quality of feature embedding. He \etal~\cite{he2020momentum} enhance the training of memory bank based contrastive learning model by storing representations from a momentum encoder instead of the trained network. Instead of storing previously computed representations, some other methods explore different instances' features within the current batch for negative sampling, and it requires a large batch size to work well~\cite{ye2019unsupervised,tian2019contrastive,chen2020simple}. 

All of the above methods require either a large batch, memory bank, or queue to provide enough negative samples for clustering or discriminating. More recently, some works have proposed advancing self-supervised pretraining without using negative samples, e.g., BYOL~\cite{grill2020bootstrap}, SimSiam~\cite{chen2020exploring}. These negative pair free self-supervised learning methods are more resilient to the changes in the batch size and more friendly to low-resource implementations.


\noindent\textbf{Data augmentation policies.}\quad 
The composition of multiple data augmentation operations is crucial in defining the contrastive prediction tasks that yield effective representations~\cite{chen2020simple,grill2020bootstrap,chen2020exploring}. Until now, most of the high-performance contrastive learning framework are designed to learn representations by maximizing agreement between differently augmented views of the same image via a contrastive loss in the feature embedding space. However, different with the supervised learning methods which can benefit from various complex data augmentation polices~\cite{chen2019destruction,devries2017improved,yun2019cutmix,cubuk2019autoaugment,cubuk2020randaugment}, there are only a few light augmentation policies playing as key contributors of the good performance of instance-wise self-supervised learning~\cite{chen2020simple,caron2020unsupervised,grill2020bootstrap,chen2020exploring}. 
The composition of the random crop, optional left-right flip, color distortion, Gaussian blur is treated as a standard and robust augmentation setting for generating augmented views of training images in unsupervised visual representation learning methods~\cite{chen2020exploring,caron2020unsupervised}. 

Our experimental study also shows that directly applying complex/heavy data augmentation policies leads to damaging performance drop or even model collapsing for negative pair free instance-wise self-supervised learning methods. These heavy data augmentations construct views with small mutual information among them. According to the infoMin principle~\cite{tschannen2019mutual,tian2020makes}, unsupervised learning methods trained on such views would result in the ``missing information'' regime of performance. Even such heavily augmented views have been demonstrated containing rich information~\cite{wang2021contrastive} but they still may mislead the feature clustering in the embedding space. 

Different with another concept of directional self-supervision loss proposed in~\cite{xu20203d} for exploiting the output consistency across different resolutions in 3D human pose estimation task, in this paper, we propose a general self-supervised learning framework DSSL for introducing various image transformations for instance-wise self-supervised learning with better theoretical justification. The contrast among heavily augmented views with strong probabilistic of missing information is disabled. Moreover, DSSL regards instance-wise self-supervised learning as optimizing a non-convex clustering task. An asymmetric loss is proposed for tightening feature clusters. As a result, DSSL can achieve stable performance improvements on various instance-wise self-supervised learning methods owing to the rich information from heavy image transformations and the data characteristic-based learning strategies.
\section{Method}
\label{sec:method}

Instance-wise self-supervised learning methods aim to learn representation by maximizing agreement among differently augmented views of the same data example in the latent visual feature space. To ease the discussion, we start by briefly summarizing the standard instance-wise self-supervised learning with a unified formulation. 

\subsection{A unified formulation}

Following the basic settings of recent works, the standard instance-wise self-supervised learning framework has four main components:
\begin{itemize}[itemsep=2pt,topsep=0pt,parsep=0pt]
\setlength\itemsep{0em}
    \item A \textit{data augmentation module} consisting with augmentation policies set $\mathcal{T}$ for generating augmented view for given image.
    \item Deep neural network \textit{encoder} $f(\cdot)$ for projecting the input images to the latent space.
    \item \textit{Projection head} $g(\cdot)$ for mapping the outputs of encoder networks to space where instance-wise self-supervised loss is applied.
    \item A \textit{self-supervised loss} function defined for an instance-wise discrimination task or a feature prediction task.
\end{itemize}

Given an input image $I$ without annotation, the data augmentation module produces augmented view pair set as 
\begin{equation}\label{eq:v_full}
    \mathbf{V}_{\mathcal{T}}=\left\{(t(I),t'(I))~|~t,t'\sim \mathcal{T}\right\}.
\end{equation}
where $t$ and $t'$ are random augmentation sampled from $\mathcal{T}$. During training, augmented view pair $(v,v')$ is sampled from $\mathbf{V}_{\mathcal{T}}$. One of the augmented views $v$ is feed into the encoders to get their visual representations $f(v)$. The projection head transforms the feature of augmented view into a vector as $z\triangleq g(f(v))$. The objective functions of instance-wise self-supervised learning is maximizing the agreement between augmented view pairs in $\mathbf{V}_{\mathcal{T}}$: $\mathcal{S}(z,y(v'))$, where $y$ works as an operation to generate the label for neural network training.

For different instance-wise self-supervised learning methods, the function of $\mathcal{S}$ and operation $y$ are implemented in different ways: 
\begin{itemize}[itemsep=2pt,topsep=0pt,parsep=0pt]
    \item In SimCLR~\cite{chen2020simple}, $y(v')=g(f(v'))$, and $\mathcal{S}$ is formulated softmax-style function across $v$, $v'$ and the negative views $v^{-}$ augmented from other images in the minbatch.
    \item In BYOL~\cite{grill2020bootstrap}, $y(v')=f_{\xi}(v')$, where $f_{\xi}(\cdot)$ is a siamese encoder with the exponential moving average weights of $f$. $\mathcal{S}$ measures the mean squared error between the normalized $z$ and $y(v')$. 
    \item In SimSiam~\cite{chen2020exploring}, $y(v')=f(v')$, but there is no gradient backward through the neural pathway computing $y(v')$. $\mathcal{S}$ measures the negative cosine similarity of given feature vector pair.
\end{itemize} 

Since the augmented view is generated randomly, above objective function is implemented as \textbf{symmetrical} form:
\begin{equation}\label{eq:UCL_loss}
    \mathcal{L}_{S} = \mathcal{S}(z,y(v')) + \mathcal{S}(z',y(v)),\\
\end{equation}
where $z'\triangleq g(f(v'))$.

\begin{figure}[!t]
  \centering
  \includegraphics[width=0.95\linewidth,page=2]{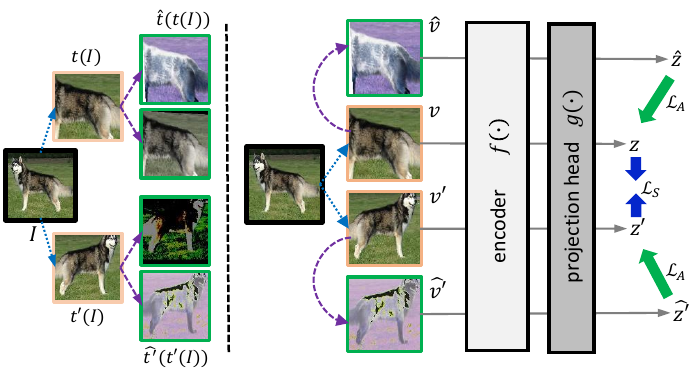}
  \caption{Illustration of our directional self-supervised learning (DSSL) framework. Left: Construction of partially ordered views. Right: Symmetric loss $\mathcal{L}_{S}$ for bidirectionally maximizing the agreement among augmented view pairs sampled from $\mathbf{V}_{\mathcal{T}}$ remains same. Asymmetric loss $\mathcal{L}_{A}$ is proposed for encouraging the representations of the heavily augmented views to be close to their source standard views, respecting the partially ordered relationship in $\mathbf{V}_{\mathcal{T}\rightarrow \widehat{\mathcal{T}}}$.}
  \label{fig:dico}
\end{figure}

\subsection{Partially-ordered views construction}
In this paper, we introduce additional heavy data augmentation policies $\widehat{\mathcal{T}}$, besides the previous carefully selected standard data augmentation polices $\mathcal{T}$. We advance the data augmentation module for generating augmented views from $\widehat{\mathcal{T}}$ and $\mathcal{T}$ jointly. In particular, a new augmented view $\widehat{v}\triangleq \widehat{t}(v)$ can be produced based on $v\triangleq t(I)$ by applying image augmentation $\widehat{t}\sim \widehat{\mathcal{T}}$, $t\sim \mathcal{T}$. It means each light augmented view $v$ would have various relevant harder augmented view $\widehat{v}$, and $\widehat{v}$ is derived from $v$.

For a well-trained unsupervised learning model, all views of the same image should be clustered closely in the visual embedding space~\cite{li2020prototypical,chen2020exploring,caron2020unsupervised}. We define the deviation of augmented view's feature $f(v)$ from the core-point (original view $I$) of its relevant cluster as $d(v)$. It is not easy to measure the deviation degree quantitatively. But, in general, larger distortion magnitudes of generating $v$ leads to larger $d(v)$. Applying additional heavy augmentation policies on a standard augmented view can generate a new view with a higher deviation from the original view. As there is no policy overlap between $\mathcal{T}$ and $\widehat{\mathcal{T}}$, and the operation of $\widehat{t}$ is non-identity. Thus, we can compare the relative magnitudes of $d$ between two views that $d(\widehat{v})>d(v)$, since $\widehat{v}$ is produced by applying additional different heavy augmentation operations $\widehat{t}$ on $v$, as shown in \cref{fig:dico} (Left).

In this way, we can construct a directed training pair collection $\mathbf{V}_{\mathcal{T}\leftarrow \widehat{\mathcal{T}}}$ from a partially ordered augmented view set as
\begin{equation}\label{eq:v_partially}
    \mathbf{V}_{\mathcal{T}\leftarrow \widehat{\mathcal{T}}}= \left\{\left(t(I),\widehat{t}(t(I))\right) |~t\sim\mathcal{T}, \widehat{t}\sim\widehat{\mathcal{T}} \right\}.
\end{equation}
After combining the original training pairs generated from $\mathcal{T}$, we get the final augmented view pairs collection for training as $\mathbf{V}_{\mathcal{T}}\cup\mathbf{V}_{\mathcal{T}\leftarrow \widehat{\mathcal{T}}}$. In particular, $\mathbf{V}_{\mathcal{T}}$ can be regarded as the collection of edges in a complete undirected graph whose vertices are the augmented views. While $\mathbf{V}_{\mathcal{T}\leftarrow \widehat{\mathcal{T}}}$ can be 
regarded as a collection of the directed edges in a directed acyclic graph, where the relations among vertices are measured by the magnitude of $d(\cdot)$.

\subsection{Directional self-supervised learning}

For every two views $v$ and $v'$ sampled from $\mathbf{V}_{\mathcal{T}}$, we sample two directed augmented view pairs $(v,\widehat{v})$, $(v',\widehat{v'})$ from $\mathbf{V}_{\mathcal{T}\leftarrow \widehat{\mathcal{T}}}$. For the partially ordered relationship between $\widehat{v}$ and $v$, we introduce a $v\leftarrow \widehat{v}$ directional asymmetric loss as
\begin{equation}
    \mathcal{L}_{A} = \mathcal{D}(\widehat{z}, y(v))+\mathcal{D}(\widehat{z'}, y(v')),
\end{equation}
where $\widehat{z}\triangleq g(f(\widehat{v}))$ and $\widehat{z'}\triangleq g(f(\widehat{v'}))$. The operation of $y(\cdot)$ stays the same with \cref{eq:UCL_loss} following the settings in standard instance-wise self-supervised learning methods, but $y(\cdot)$ is only computed over the standard views in $\mathcal{L}_{A}$. This makes the self-supervised learning to be directional and asymmetrical. It means the optimization objective of $\mathcal{L}_{A}$ is to force the representation of heavily augmented view close to the representation of its relevant source view. $\mathcal{D}$ simply measure the negative cosine similarity among two vectors:
\begin{equation}
    \mathcal{D}(z,y) = -\frac{\left \langle z,y \right \rangle}{\left \| z\right \|_2 \left \| y\right \|_2}.\nonumber
\end{equation}

As shown in \cref{fig:dico} (Right), the objective function of our proposed directional self-supervised learning for given two standard views and their relevant harder views can be expressed as:
\begin{equation}\label{eq:dico}
    \mathcal{L}_{DSSL} = \mathcal{L}_{S} + \mathcal{L}_{A}.
\end{equation}
The total loss of DSSL can be measured by $\sum_{\mathbf{V}_{\mathcal{T}}}\mathcal{L}_{S}+\sum_{\mathbf{V}_{\mathcal{T}\leftarrow \widehat{\mathcal{T}}}}\mathcal{L}_{A}$ averaged over all images. DSSL can be easily implemented on various instance-wise self-supervised learning frameworks in only a few lines of Pseudocode. \cref{alg} shows the Pseudocode applying DSSL for unsupervised learning framework SimSiam~\cite{chen2020exploring}. 

\begin{lrbox}{\mycode}
\begin{lstlisting}[language=Python, escapeinside={(*}{*)}]
# f: feature encoder; g: prediction head;
for I in dataloader:
    (*$v$*), (*$v'$*) = (*$t$*)(I), (*$t'$*)(I)  # standard data augmentation
    (*$\widehat{v}$*), (*$\widehat{v'}$*) = (*$\widehat{t}$*)(I), (*$\widehat{t'}$*)(I)  # heavy data augmentation
    (*$y$*), (*$y'$*), (*$\widehat{y}$*), (*$\widehat{y'}$*) = f((*$v$*)), f((*$v'$*)), f((*$\widehat{v}$*)), f((*$\widehat{v'}$*))
    (*$z$*), (*$z'$*), (*$\widehat{z}$*), (*$\widehat{z'}$*) = g((*$y$*)), g((*$y'$*)), g((*$\widehat{y}$*)), g((*$\widehat{y'}$*))
    # loss computation
    L = D((*$z$*), (*$y'$*))/4 + D((*$z'$*), (*$y$*))/4 + D((*$\widehat{z}$*), (*$y$*))/4 + D((*$\widehat{z'}$*), (*$y'$*))/4
    
    L.backward()  # back-propagate
    update(f, h)  # SGD update
    
def D(z, y):  # negative cosine similarity
    y = y.detach()  # stop gradient
    # l2-normalize
    z = normalize(z, dim=1)
    y = normalize(y, dim=1)
    return -(z * y).sum(dim=1).mean()
\end{lstlisting}
\end{lrbox}

\begin{algorithm}[t]
\usebox{\mycode}
\caption{DSSL for SimSiam Pseudocode, PyTorch-like}
 \label{alg}
\end{algorithm}

\noindent\textbf{Relation to previous self-supervised learning methods.}\quad  We formulate the general self-supervised learning that considering all combinations of views augmented by $\mathcal{T}$ and $\widehat{\mathcal{T}}$ jointly as:
\begin{equation}
 \alpha\sum_{\mathbf{V}_{\mathcal{T}}}\mathcal{L}_{S}+\beta\sum_{\mathbf{V}_{\widehat{\mathcal{T}}}}\mathcal{L}_{S} + \gamma\sum_{\mathbf{V}_{\mathcal{T}\leftarrow \widehat{\mathcal{T}}}}\mathcal{L}_{A} + \delta\sum_{\mathbf{V}_{\widehat{\mathcal{T}}\leftarrow \mathcal{T}}}\mathcal{L}_{A},
 \label{eq:general_formulation}
\end{equation}
where $\mathbf{V}_{\widehat{\mathcal{T}}}$ and $\mathbf{V}_{\widehat{\mathcal{T}}\leftarrow \mathcal{T}}$ are the augmented view collections constructed according to \cref{eq:v_full} and \cref{eq:v_partially} respectively. $\leftarrow$ indicates the direction of predicting or contrasting. The $\alpha$, $\beta$, $\gamma$ and $\delta$ are the weights for each part of the loss. Previous self-supervised learning methods fairly treat all augmented views, thus the objective functions of these methods can be regards as the situation that $\alpha=\gamma=\beta=\delta=1$. While the loss function of DSSL ($\mathcal{L}_{DSSL}$) can be regarded as the situation with $\alpha=\gamma=1$, $\beta=\delta=0$. 

Considering that the view augmented from additional heavy augmentation policies $\widehat{\mathcal{T}}$ would tend to lose more information than its source view augmented from $\mathcal{T}$, a natural solution is to fit a lower confidence score for heavily augmented views when they play as the targets for optimizing the self-supervised loss. Thus, an ideal settings of the loss weights should ensure $\alpha,\gamma > \beta,\delta$. Further, we designed various self-supervised learning paradigms in \cref{ablation} to analyze the positive/negative impact of each components in \cref{eq:general_formulation} and demonstrate the necessity of asymmetric loss for heavy augmentations in \cref{eq:dico}.

\section{Experimental Results}\label{sec:exp}
In this section, we empirically study the DSSL behaviors by introducing various heavy data augmentation policies.

\subsection{Implementation details}
\textbf{Standard augmentations.}\quad We use the same sequence of image augmentations as in previous instance-wise self-supervised learning methods~\cite{chen2020exploring,chen2020simple,grill2020bootstrap}, including random \textit{cropping and resizing}, \textit{horizontal flip}, \textit{color distortion}, \textit{grayscale} and \textit{Gaussian blur}. Each of the above augmentations has been proved effective in at least one typical instance-wise self-supervised learning method. The composition of all above image augmentations are treated as the standard augmentation $\mathcal{T}$.

\noindent\textbf{Heavy augmentations.}\quad Inspired by the settings in InfoMin~\cite{tian2020makes}, we use \textit{RandAugment}~\cite{cubuk2020randaugment} and \textit{Jigsaw}~\cite{noroozi2016unsupervised,chen2019destruction} as heavy augmentations $\widehat{\mathcal{T}}$. These two augmentations have proven effective for supervised representation learning, and negative pairs required instance-wise contrastive learning. But we find that these augmentations result in lousy performance or even model collapsing for the negative-pairs-free unsupervised learning. We denote \textit{RandAugment} as \textit{RA}($n,m$), where $n$ is the number of augmentation policies randomly selected from 14 predefined augmentations, $m$ is the magnitude for all the transformation. Unless otherwise specified, we equip a sequential combination of \textit{RA}(2,5) and \textit{Jigsaw} with $4\!\times\!4$ grids as heavy augmentation for experimental results reported in this paper. 



\noindent\textbf{Compared methods.}\quad We compare three typical self-supervised learning frameworks, including SimSiam, BYOL, and SimCLR, as shown in Table~\ref{tab:compared_methods}. SimCLR uses the normalized temperature-scaled cross entropy as $\mathcal{S}$, and the comparisons among positive and negative views are required for model training. Both BYOL and SimSiam are negative sample free methods, and they stop the gradient of the neural branch for computing the label $y$, except BYOL that applies a momentum encoder for updating $y$.
\begin{table}[t]
\renewcommand{\arraystretch}{1.2}
\small
\setlength{\tabcolsep}{1.2mm}{
\begin{tabular}{m{1.6cm}|m{1cm}<{\centering}|m{1.1cm}<{\centering}|m{1.4cm}<{\centering}|m{1cm}<{\centering}|m{0.8cm}<{\centering}}
Method  & negative pairs & stop gradient & momentum encoder & reported & repro.\\
\Xhline{2\arrayrulewidth}
SimCLR~\cite{chen2020simple}  & \checkmark & & & 60.1 & 59.5\\
SimSiam~\cite{chen2020exploring}  & & \checkmark &   & 67.7 & 67.4  \\
BYOL~\cite{grill2020bootstrap}  & & \checkmark & \checkmark & 66.5 & 67.8  \\
\end{tabular}}
\caption{Three instance-wise self-supervised learning methods used for comparisons and analysis in this paper.}
\label{tab:compared_methods}
\end{table}

For a fair comparison, we generated two augmented view pairs per image each time for negative sample free methods (SimSiam and BYOL) to keep the number of feature prediction pairs the same with their corresponding \textit{w/ DSSL} versions. Moreover, for SimCLR that considers negative samples in the whole min-batch (batch size of $n$), its \textit{w/ DSSL} version only adds one additional heavily augmented view with asymmetric loss item for each sample. The computation slightly increases from $2n^2$ to $2n^2\!+\!2n$ for each min-batch after applying DSSL on SimCLR. The number of feedforward and backward passes remains the same. More implementation details of compared methods and DSSL can be found in the Appendix.

\begin{figure*}[ht]
\begin{minipage}[t!]{.62\linewidth}
\small
\centering
\renewcommand{\arraystretch}{1.15}
\begin{tabular}{l|ccc|ccc}
\multirow{2}{*}{Method} & \multicolumn{3}{c|}{CIFAR-10} & \multicolumn{3}{c}{ImageNet} \\
 & SimCLR & BYOL & SimSiam & SimCLR & BYOL & SimSiam\\
\Xhline{2\arrayrulewidth}
repro. & 91.5 & 92.2 & 92.1 & 59.5 & 67.8 & 67.4 \\
w/ $\widehat{\mathcal{T}}$ & 92.2 & 94.4 & \textit{collapse} & 60.0 & \textbf{68.3} & \textit{collapse} \\\hline
w/ DSSL & \textbf{93.2} & \textbf{94.7} & \textbf{94.5} & \textbf{60.3} & \textbf{68.3} & \textbf{68.6} \\
\end{tabular}
\captionof{table}{Comparisons on linear evaluation accuracies (\%). \textit{repro}: our reproduction of each method. \textit{collapse}: model collapsed during training. \textit{w/ $\widehat{\mathcal{T}}$}: training views are jointly augmented from standard and heavy augmentations. Heavy augmentations perform unstably even model collapsing, while DSSL \textit{consistently} benefits from $\widehat{\mathcal{T}}$.}
\label{tab:imagenet_cifar_exp}
\end{minipage}
~~~
\begin{minipage}[t!]{.37\linewidth}
  \centering
  \includegraphics[width=0.87\linewidth,page=1]{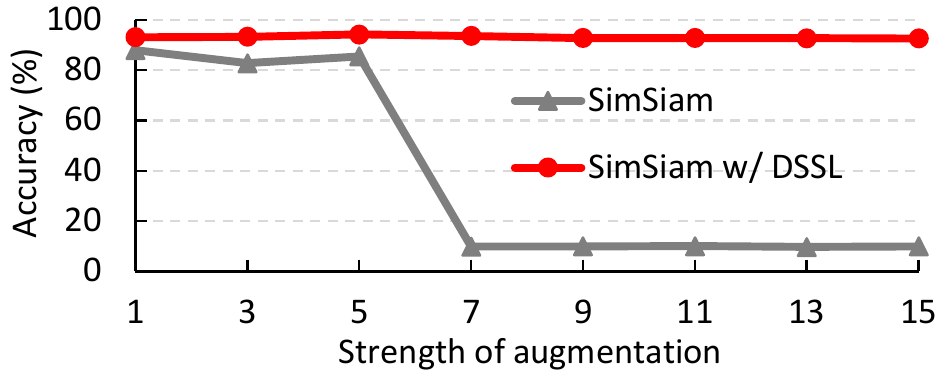}
  \caption{Performance comparisons on CIFAR-10 linear evaluation accuracy for SimSiam by applying heavy augmentation \textit{RA}($n=2$) across varying distortion magnitudes $m$.}
  \label{fig:magnitudes}
\end{minipage}%
\end{figure*}

\noindent\textbf{Training details.}\quad
Following the practice of the previous self-supervised learning methods, we use ResNet-50 and ResNet-18 as the basic feature encoder $f$ for the experiments on ImageNet ILSVRC-2012~\cite{russakovsky2015imagenet} and CIFAR-10~\cite{krizhevsky2009learning} dataset, respectively. We strictly follow the network architecture of projection head $g$, initialization, optimizer represented in these methods' original paper. Except that we apply the same SGD optimizer and learning rate schedule of SimSiam to BYOL since it can slightly improve the performance of BYOL. Existing works differ considerably in batch size and training epochs, which could significantly influence performance. We, therefore, compare all models of the same batch size of 512 and training epochs of 100 for unsupervised training on ImageNet, which is a resource-friendly implementation. 

We elaborate the implementation details of the compared methods below. 
\begin{itemize}[itemsep=2pt,topsep=0pt,parsep=0pt]
    \item \textbf{SimCLR}: We used the PyTorch repo\footnote{https://github.com/AndrewAtanov/simclr-pytorch} officially recommended by the authors.
    \item \textbf{BYOL}: Our re-implemented BYOL has higher linear evaluation accuracy than the BYOL (100ep) results of ImageNet linear evaluation reported in the SimSiam paper (67.8\% vs. 66.5\%).
    \item \textbf{SimSiam}: To align settings of all compared methods in this paper, we used \textit{lr} with cosine decay but did not fix the \textit{lr} of prediction MLP during linear evaluation. The result of our reproduction is 67.4\% on ImageNet (vs. 67.7\% reported in the original paper). This performance gap is acceptable.
\end{itemize}

\subsection{Main results and discussions}
\textbf{Linear evaluation on ImageNet.}\quad We apply linear evaluation to measure the quality of the representations of DSSL based models after self-supervised pretraining on the unlabelled training images of the ImageNet dataset. Specifically, we train a linear classifier on top of the pre-trained representation. During training, the parameters of the backbone network (feature encoder) are frozen, while only the last fully connected layer is updated via backpropagation.

\cref{tab:imagenet_cifar_exp} reports the top-1 accuracy of compared methods and their DSSL versions. 
Same with the experimental conclusion of InfoMin~\cite{tian2020makes}, the evident advantages of variant information in the harder views augmented by $\widehat{\mathcal{T}}$ can further benefit SimCLR, which is a contrastive learning method relying on negative pairs. Maximizing the dissimilarity among different instances is a direct and effective way to learn a well-separated visual embedding space. Such a manner is also robust to the missing-information hard views. However, for the negative pairs free method SimSiam, the heavy augmentations usually result in the model collapsing. Momentum encoder can increase model training stability since momentum would neutralize the misleading information from inconsistent representations. BYOL is more robust to the hard views than SimSiam. We further analyse the robustness of the momentum encoder in the later subsections.

Our DSSL prevents closing the similarity among heavily augmented views since the commonality among these views is usually scarce, and forcing assimilating them would lead to the model collapsing. 
As a result, DSSL can make these views no longer risky for negative pairs free self-supervised learning methods and make them play a bigger role in contrastive learning methods which relying on negative pairs. 

Although the improvement of DSSL on BYOL is limited under this setting, such limitation is easy to break after introducing more heavy augmentations, as we listed in \cref{tab:byol_limiations}. After equipping \textit{RA}(8,16) and \textit{UA}~\cite{lingchen2020uniformaugment}, DSSL can significantly boost the accuracy of BYOL linear evaluation +7.2\%, +2.2\% on ImageNet, respectively. DSSL further taps potentials in various heavy augmentations and results in stable performance improvement.   

\noindent\textbf{Linear evaluation on CIFAR-10.}\quad 
Similar to the implementation in the ImageNet dataset, we use the unlabeled training images in CIFAR-10 for self-supervised learning based on the ResNet-18 backbone. We follow the standard settings used in the CIFAR experiments of SimSiam~\cite{chen2020exploring} that SGD with a learning rate 0.03, cosine $lr$ decay schedule for 800 epochs, image size of 32$\times$32, and a batch size of 512. We first train the ResNet-18 feature encoder on unlabeled CIFAR-10 images and then freeze the backbone to train a linear tasks-specific head on CIFAR-10 dataset with annotations. 

\cref{tab:imagenet_cifar_exp} reports the top-1 accuracy of linear classifier trained on CIFAR-10. Our reproductions of SimSiam (92.1) and SimCLR (91.5) have higher performance than that reported in the paper of SimSiam under the same settings (91.8, 91.1). 
Similar to the experimental results in ImageNet, 
after applying DSSL, all three unsupervised learning methods \textbf{outperform the supervised ResNet-18}, whose top-1 accuracy is 93.02\%.

\begin{figure*}[htbp]
  \centering
  \includegraphics[width=0.97\linewidth,page=2]{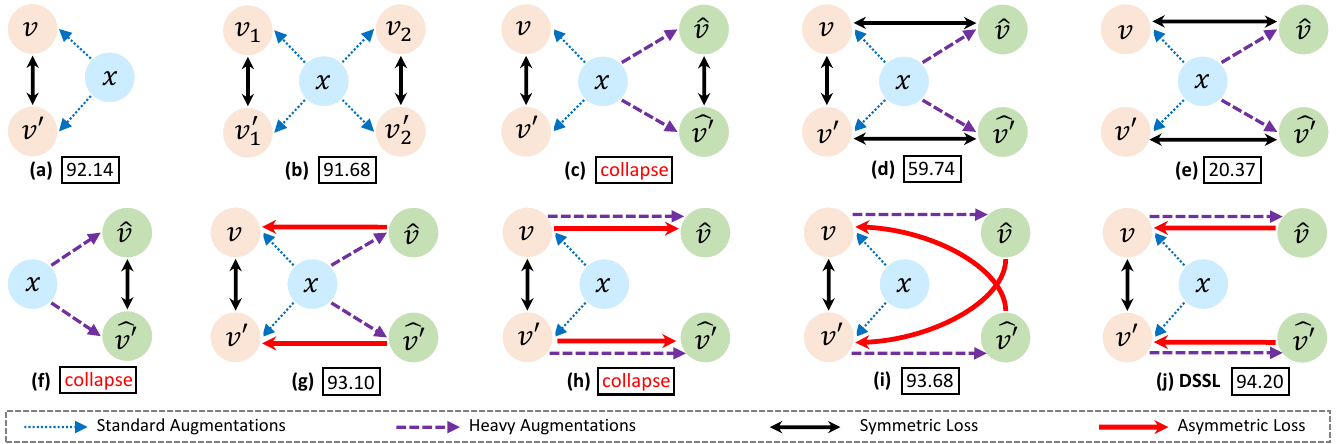}
  \caption{Comparisons of different views construction and instance-wise self-supervised learning mechanisms. Top-1 accuracies (\%) of CIFAR-10 linear classifiers trained on the freeze representations are listed in the box below.}
  \label{fig:justification}
\end{figure*}

\noindent\textbf{Impact of distortion magnitudes of $\widehat{\mathcal{T}}$.}\quad In \cref{fig:magnitudes}, we compared the results from the standard SimSiam and its DSSL version by introducing heavy augmentation with various distortion magnitudes. We use the hyperparameter $m$ of RandAugment to control the distortion magnitudes of the randomly selected policies. It can be found that, after increasing $m\ge5$, the self-supervised models result in divergence. In contrast, SimSiam \textit{w/ DSSL} can converge stably in a wide range of $[1,9]$. It further demonstrates the robustness of our proposed directional self-supervised learning on partially ordered views.

\begin{table}[t]
\renewcommand{\arraystretch}{1.15}
\centering
\small
\begin{tabular}{l|cc|cc}
Method & \multicolumn{2}{c|}{CIFAR-10} & \multicolumn{2}{c}{ImageNet} \\
\Xhline{2\arrayrulewidth}
BYOL (repro.) & \multicolumn{2}{c|}{92.2} & \multicolumn{2}{c}{67.8} \\
\hline
$+\widehat{\mathcal{T}}$: & \textit{RA}(4,10) & \textit{RA}(8,16) & \textit{RA}(8,16) & \textit{UA} \\
\hline
BYOL w/ $\widehat{\mathcal{T}}$ &  84.1 & \textit{collapse} & 60.6 & 67.1\\
BYOL w/ DSSL & \textbf{94.4} & \textbf{94.0} & \textbf{67.8} & \textbf{69.3}\\
\end{tabular}
\caption{Linear evaluation accuracies (\%) of BYOL by applying more heavy augmentations ($+\widehat{\mathcal{T}}$).}
\label{tab:byol_limiations}
\end{table}

\noindent\textbf{Robustness analysis of momentum encoder.} As shown in \cref{tab:imagenet_cifar_exp}, the momentum encoder based method BYOL represents better robustness to the heavy augmentations than SimSiam. To study the limitation of negative sample free self-supervised methods and look into the boundary of momentum updating for the encoder, we further strengthen the magnitudes of heavy augmentations for BYOL. In particular, \textit{UA} and \textit{RA} with more augmentation policies and higher distortion magnitudes are selected. The results presented in \cref{tab:byol_limiations} support our hypothesis: the heavy augmentations play as a ``double-edged sword'' for instance-wise self-supervised learning; without using negative pairs, the model would be easily misled by the heavily augmented views. Although momentum encoder can ease heavily augmented views' side effects up to a point, DSSL is a more fundamental schema for handling various image transformations during instance-wise self-supervised learning.

\noindent\textbf{Transfer to other vision tasks.}\quad We evaluate the representations benefited from DSSL on different tasks relevant to computer vision practitioners, including COCO~\cite{lin2014microsoft} object detection and instance segmentation. Unlike linear evaluation, we fine-tune the 100-epoch pre-trained BYOL and SimSiam models end-to-end in the COCO dataset. We apply the public codebase Detectron~\cite{Detectron2018} to implement Mask R-CNN~\cite{he2017mask} detector and evaluate the COCO 2017 val. \cref{tab:coco_exp} shows that 
DSSL can consistently improve the standard self-supervised learning method on downstream object detection and instance segmentation tasks.

\begin{table}[t]
\renewcommand{\arraystretch}{1.2}
\small
\setlength{\tabcolsep}{1.5mm}{
\begin{tabular}{l|ccc|ccc}
\multirow{2}{*}{pre-train} & \multicolumn{3}{c|}{COCO detection} &  \multicolumn{3}{c}{COCO segmentation}\\
& AP$_{50}$ & AP & AP$_{75}$ & AP$^{m}_{50}$ & AP$^{m}$ & AP$^{m}_{75}$ \\
\Xhline{2\arrayrulewidth}
ImageNet sup. & 59.2 & 37.7 & 40.9 & 55.8 & 33.9 & 35.8 \\
\hline
BYOL (repro.) & 56.9 & 36.9 & 40.2 & 54.1 & 34.0 & 36.5 \\
BYOL w/ DSSL & \textbf{57.8} & \textbf{37.6} & \textbf{40.8} & \textbf{55.0} & \textbf{34.6} & \textbf{37.0} \\
\hline
SimSiam$^\dagger$ & 57.5 & 37.9 & 40.9 & 54.2 & 33.2  & 35.2 \\
SimSiam (repro.) & 58.0 & 37.7 & 40.9 & 55.0 &34.7  & 37.1 \\
SimSiam w/ DSSL & \textbf{58.2} & \textbf{38.1} & \textbf{41.5}  &\textbf{55.5} &\textbf{35.1}  & \textbf{37.6} \\
\end{tabular}}
\caption{Results of object detection and instance segmentation fine-tuned on COCO. We adopt Mask R-CNN R50-FPN with a 1x schedule and report the bounding box AP and mask AP on COCO 2017 val. Our reproduction and DSSL versions are based on 100-epoch pre-training in ImageNet. $^\dagger$: the reported 200-epoch result.}
\label{tab:coco_exp}
\end{table}

\subsection{Justification and ablation study}\label{ablation}
To understand how the asymmetric loss and partially-ordered views impact the representations learning, we design experiments with various settings of instance-wise self-supervised learning on the CIFAR-10 dataset. We apply the random resize-crop with color distortion as the standard image transformation, and apply \textit{RA}(2,5) as heavy augmentation. \cref{fig:justification} shows the \textit{augmentation, loss function design} v.s. \textit{linear evaluation accuracy}. 

Obviously, 
directly making all views to be similar without distinguishing standard and heavily augmented views hurts the performances (\textbf{c}:\textcolor{red}{\textit{collapse}}, \textbf{d}:59.7\textcolor{red}{$\downarrow$}, \textbf{e}:20.4\textcolor{red}{$\downarrow$}). According to the hypothesis proposed in InfoMin~\cite{tian2020makes}, it is the information about the task-relevant variable that is discarded by the hard view, degrading performance. Further, we study the impact of each components in Eq.~(\ref{eq:general_formulation}) and find below four phenomenons effectively illustrate the rationality of our proposed DSSL (Eq.~\ref{eq:dico}).

(i) Introducing symmetric loss $\sum_{\mathbf{V}_{\widehat{\mathcal{T}}}}\mathcal{L}_{S}$ among heavily augmented views results in collapsing during model training (\textbf{c}, \textbf{f}).

(ii) \textit{Fairly contrasting between the standard view and heavily augmented view has negative effects}. After comparing the results of (\textbf{h}:\textcolor{red}{\textit{collapse}}) and (\textbf{j}:94.2\textcolor{green}{$\uparrow$}), the misleading impact of $\sum_{\mathbf{V}_{\widehat{\mathcal{T}}\leftarrow \mathcal{T}}}\mathcal{L}_{A}$ can be observed obviously.

(iii) \textit{Equipping directional feature prediction between heavily augmented view and standard view results in stable performance improvement} (\textbf{d}:59.7 vs. \textbf{g}:93.1).

(iv) \textit{Partially-ordered views construction mechanism results in better performance}, as we compared among (\textbf{g}:93.1, \textbf{i}:93.7, \textbf{j}:94.2:\textcolor{green}{$\uparrow$}). It mainly due to that the mutual information between $v$ and its derived harder view $\widehat{v}$ is guaranteed up to a point. Such a mechanism can prevent the issue of unexpected missing information. 

\begin{figure}[t]
  \centering
  \includegraphics[width=0.72\linewidth,page=1]{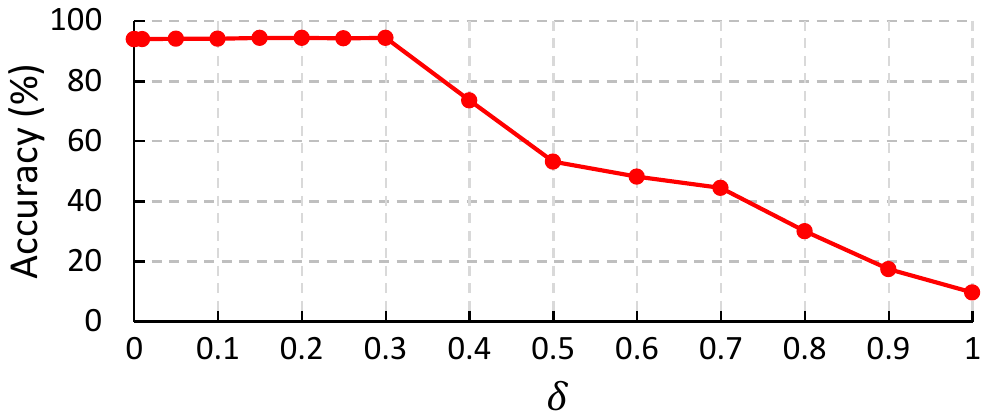}
  \caption{CIFAR-10 linear evaluation accuracy of SimSiam optimized according to \cref{eq:general_formulation} by setting $\alpha=1, \beta=0$, and $\lambda=1-\delta$ across varying $\delta$. \textit{RA} is applied to construct hard views.}
  \label{fig:soft-weighted}
\end{figure}

For more critical analysis on the influence of $\sum_{\mathbf{V}_{\widehat{\mathcal{T}}\leftarrow \mathcal{T}}}\mathcal{L}_{A}$, we trained a SimSiam model according to \cref{eq:general_formulation} by setting $\lambda+\delta=1, \alpha=1, \beta=0$. As shown in \cref{fig:soft-weighted}, the negative impact of mapping heavily augmented view's feature to standard view only emerges after the value of $\delta$ raised to a threshold. 
At the early stage, the linear evaluation accuracy remains stable. This phenomenon further reveals mapping standard view's feature to heavily augmented view would be high-risk and low-return. 

Moreover, \cref{eq:general_formulation} is the theoretical formulation of the standard instance-wise self-supervised learning method while fairly treating all views. Our DSSL only activates two training view pairs (comparing \cref{eq:general_formulation} and \cref{eq:dico}). Thus DSSL's computation complexity can be thought to be only half of the previous methods when considering all possible view pairs during training.

\section{Conclusion and Discussion}

We propose a directional self-supervised learning (DSSL) framework for unsupervised visual representation learning. Compared to standard self-supervised learning methods, our proposed framework benefits from more heavy image transformations and results in stable performance improvement on various vision tasks. Moreover, DSSL is easy to implement and compatible with most of the typical instance-wise self-supervised learning methods. The core concepts of DSSL can further guide the loss design of self-supervised learning. According to our formulation in Eq.~(\ref{eq:general_formulation}), the soft weighted version of DSSL respecting to view characteristics is also worthy of further exploration.

{\small
\bibliographystyle{ieee_fullname}
\bibliography{egbib}
}

\appendix
\begin{figure*}[t]
  \centering
  \includegraphics[width=1\linewidth,page=1]{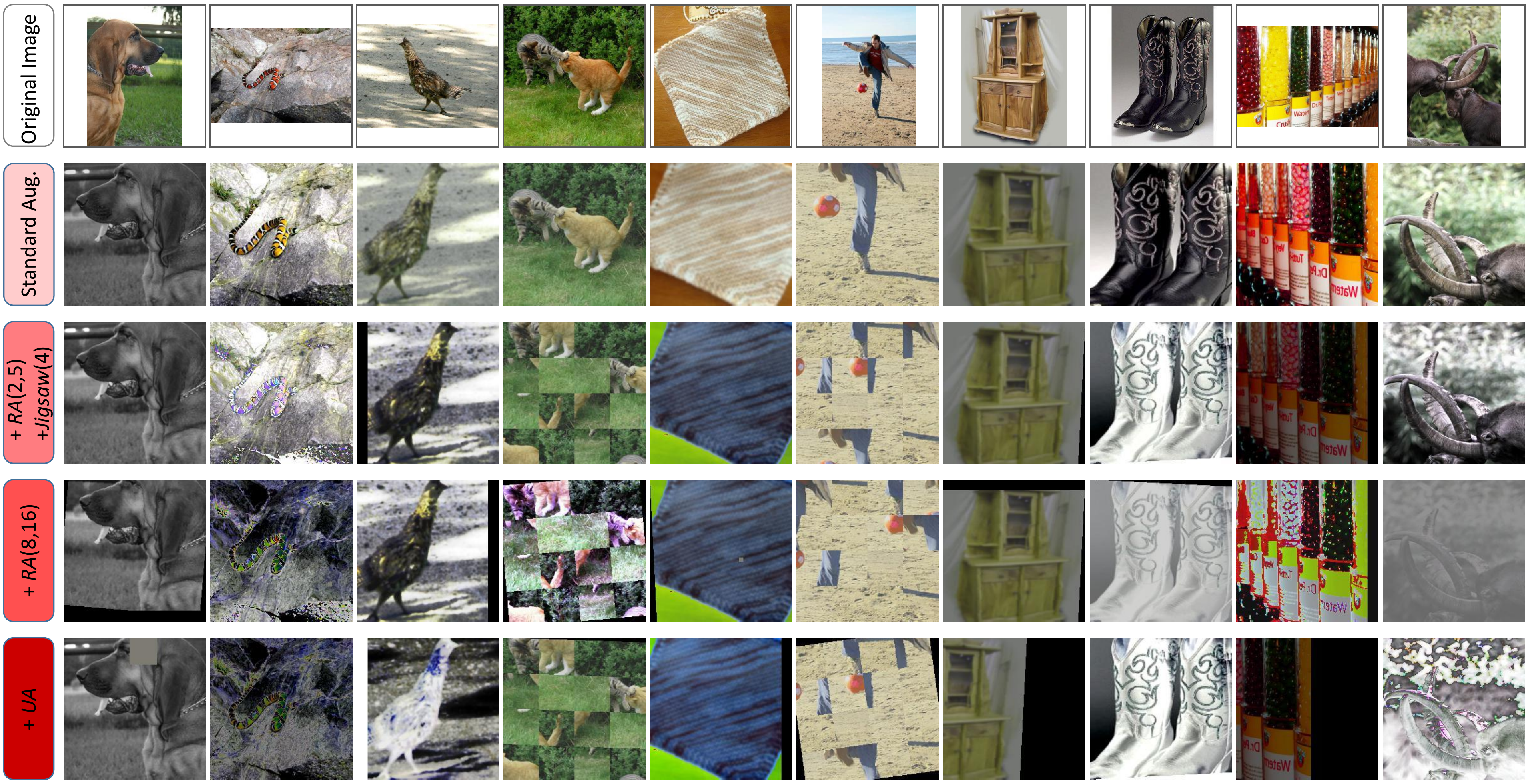}
  \caption{Example views augmented by various data augmentation on ImageNet dataset. For each original image (in the first row), we show their standard views (in the second row), the heavily augmented views derived from the standard views by applying $\widehat{\mathcal{T}}$ as \textit{RA}(2,5) and \textit{Jigsaw}(4) (in the third row), and the results of the additional heavy augmentation \textit{RA}(8,16) and \textit{UA} (in the last two rows).}
  \label{fig:examplesimages}
\end{figure*}
\section{Implementation details}
\subsection{Standard augmentations} First, an 224$\times$224 image patch is randomly \textit{cropped} from the image of \textit{scale} in [0,2, 1.0] with random \textit{horizontal flip}, followed by a \textit{color distortion} with brightness, contrast, saturation, hue strength of \{0.4, 0.4, 0.4, 0.1\} with an applying probability of 0.8, and an optional \textit{grayscale} conversion with applying probability of 0.2. At last, \textit{Gaussian blur} with kernel std in [0.1, 2.0] processes the image patch with applying probability of 0.5. Each of above image augmentations has been proved to be effective in at least one typical instance-wise self-supervised learning method. The composition of all above image augmentations are treated as standard augmentation $\mathcal{T}$.

\subsection{Heavy augmentations} \textit{RandAugment} covers various of image transformations and has been demonstrated with significant performance improvement on supervised representation learning. It has two tunable hyperparameters $m,n$, where $n$ is the number of augmentation policies randomly selected from 14 predefined augmentations, $m$ is the magnitude for all the transformation. Larger $m$ and $n$ results in stronger image transformation. In this paper, we denote \textit{RandAugment} as \textit{RA}($n,m$). Similarity, \textit{Jigsaw} has also been  used in pretext task based self-supervised representation learning and fine-grained visual patterns. The hyperparameter $n$ for \textit{Jigsaw} is the number of grid $n\times n$ for each images. Our heavy augmentations $\widehat{\mathcal{T}}$ consists of \textit{RA}(2,5) and \textit{Jigsaw}(4) with probability of 0.9 and 0.1 respectively. Except these two augmentation policies, we applied  \textit{RA}(8,16) and \textit{UniformAugment} as the additional heavy augmentation ($+\widehat{\mathcal{T}}$) for robustness analysis experiments in Table 3. There is no hyperparameter for \textit{UniformAugment}. It is a search-free DA method consists of 13 different image transformations by assuming that the augmentation space is approximately distribution invariant.

We show some example images augmented by the standard augmentations and various heavy augmentations in \Cref{fig:examplesimages}. All heavy augmented images are derived from the standard augmentation results. 


\subsection{ImageNet linear evaluation} For linear evaluation on ImageNet, we follow a similar procedure of SimCLR, BYOL, and SimSiam. The frozen ResNet's global average pooling layer outputs are used to train a supervised linear classifier. For SimCLR, we use the SGD optimizer as the original paper recommended. For SimSiam and BYOL, we used a LARS optimizer. We trained all linear classifier with base $lr=1.6$ (following $lr$ = 0.1 $\times$ BatchSize/256) and batch size of 4096 for 90 epochs. For the SGD optimizer, we decay the learning rate with the linear decay schedule. For the LARS optimizer, we decay the learning rate with the cosine decay schedule. After training the linear classifiers, we evaluate them on the center cropped 224$\times$224 resolution inputs in validation set.

\begin{lrbox}{\mycode}
\begin{lstlisting}[language=Python, escapeinside={(*}{*)}]
# f: feature encoder; g: prediction head;
# aug: data augmentations;

for I in dataloader:
    # random augmentation
    (*$v_1$*), (*$v_1'$*), (*$v_2$*), (*$v_2'$*) = aug(I), aug(I), aug(I), aug(I) 
    # projections
    (*$y_1$*), (*$y_1'$*), (*$y_2$*), (*$y_2'$*) = f((*$v_1$*)), f((*$v_1'$*)), f((*$v_2$*)), f((*$v_2'$*))
    # predictions
    (*$z_1$*), (*$z_1'$*), (*$z_2$*), (*$z_2'$*) = g((*$y_1$*)), g((*$y_1'$*)), g((*$y_2$*)), g((*$y_2'$*))
    # loss
    L = D((*$z_1$*),(*$y_1'$*))/4 + D((*$z_1'$*),(*$y_1$*))/4 + D((*$z_2$*),(*$y_2'$*))/4 + D((*$z_2'$*),(*$y_2$*))/4
    
    L.backward()  # back-propagate
    update(f, h)  # SGD update
    
def D(z, y):  # negative cosine similarity
    y = y.detach()  # stop gradient
    # l2-normalize
    z = normalize(z, dim=1)
    y = normalize(y, dim=1)
    return -(z * y).sum(dim=1).mean()
\end{lstlisting}
\end{lrbox}

\begin{algorithm}[t]
\usebox{\mycode}
\caption{The Pseudocode of SimSiam baseline with 2 view-pairs}
 \label{alg_twoview}
\end{algorithm}

\section{Compared methods} 
%
\cref{alg_twoview} shows the Pseudocode of our re-implemented SimSiam with four views. We list the SimSiam results of its original version (with two standard views), our re-implemented version with four standard views, and re-implemented versions with four heavily augmented views in \Cref{tab:repro}. Although previous work SwAV~\cite{caron2020unsupervised} shows that contrasting more image crops with smaller size (e.g., 96$\times$96) with regular image crops (224$\times$224) results in performance improvement owing to more visual details in low-resolution crops are highlighted, introducing more views with the same resolution in each min-batch has little impact on performance. Even though the number of views is increased, the frequency of model updating during training and the number of raw samples in min-batch keep the same and results in similar results (one view-pair vs. two view-pairs).

\begin{table}[t]
\renewcommand{\arraystretch}{1.2}
\centering
\small
\begin{tabular}{l|c|c}
Baseline Implementation & CIFAR-10 & ImageNet\\
\Xhline{2\arrayrulewidth}
1 standard view-pair  & 92.14 & 67.7 \\
2 standard view-pairs & 91.68  & 67.4 \\
2 heavily augmented view-pairs & \textit{collapse} & \textit{collapse} \\
\end{tabular}
\caption{Linear evaluation Top-1 accuracy of SimSiam on ImageNet and CIFAR-10 with same training epochs but different training pair settings.}
\label{tab:repro}
\end{table}

For SimCLR that considers negative samples in the whole min-batch (batch size of $n$), DSSL only add two heavily augmented views and two $t(I)\!\leftarrow\!\widehat{t}(I)$ asymmetric loss items for each image. The computation complexity slightly increases from $2n^2$ to $2n^2\!+\!2n$ for each min-batch after applying DSSL.

As shown in Algorithm 1 and Eq. (5), DSSL has no additional hyperparameters comparing to the standard instance-wise self-supervised learning methods.



\end{document}